 \documentclass[final,3p,twocolumn,times]{elsarticle}

\usepackage{amssymb}

\journal{}

\usepackage{array}
\usepackage{makecell}

\usepackage{caption}
\usepackage{longtable}
\usepackage{color}
\usepackage{subfigure}
\usepackage{array}

\usepackage{floatrow}
\usepackage{pgfplots} 
\newfloatcommand{capbtabbox}{table}[][\FBwidth]

\usepackage{booktabs}
\usepackage{xcolor}
\usepackage{color, colortbl}
\definecolor{Gray}{gray}{0.9}
\definecolor{LightCyan}{rgb}{0.92,1,1}
\usepackage{multirow}
\usepackage{amsmath}
\begin{document}

\begin{frontmatter}


\title{A Probabilistic-Logic based Commonsense Representation Framework for Modelling Inferences with Multiple Antecedents and Varying Likelihoods}


\author{Shantanu Jaiswal}

\affiliation{organization={Social and Cognitive Computing Department, Institute of High Performance Computing, A*STAR},
            country={Singapore}
            }

\author{Liu Yan}
\author{Dongkyu Choi}
\author{Kenneth Kwok}


\begin{abstract}

Commonsense knowledge-graphs (CKGs) are important resources towards building machines that can `reason' on text or environmental inputs and make inferences beyond perception.
While current CKGs encode world knowledge for a large number of concepts and have been effectively utilized for incorporating commonsense in neural models
, they primarily encode declarative or single-condition inferential knowledge and assume all conceptual beliefs to have the same likelihood. 
Further, these CKGs utilize a limited set of relations shared across concepts and lack a coherent knowledge organization structure resulting in redundancies as well as sparsity across the larger knowledge graph. Consequently, today's CKGs, while useful for a first level of reasoning, do not adequately capture deeper human-level commonsense inferences which can be more nuanced and influenced by multiple contextual or situational factors. 

Accordingly, in this work, we study how commonsense knowledge can be better represented by -- (i) utilizing a probabilistic logic representation scheme to model composite inferential knowledge and represent conceptual beliefs with varying likelihoods, and (ii) incorporating a hierarchical conceptual ontology to identify salient concept-relevant relations and organize beliefs at different conceptual levels. Our resulting knowledge representation framework can encode a wider variety of world knowledge and represent beliefs flexibly using grounded concepts as well as free-text phrases. 
As a result, the framework can be utilized as both a traditional free-text knowledge graph and a grounded logic-based inference system more suitable for neuro-symbolic applications.
We describe how we extend the PrimeNet knowledge base with our framework 
through crowd-sourcing and expert-annotation, and demonstrate how it can be applied for deeper and more interpretable passage-based semantic parsing and question answering.

\end{abstract}




\end{frontmatter}


\section{Introduction}
\label{sec:introd}


The development of commonsense knowledge resources has been a prominent line of research in enabling artificial intelligence (AI) models to not only interpret text or environmental scenes, but to also derive inferences upon them to support higher-level cognitive tasks~\cite{davis2015commonsense,camcom}. These knowledge-bases are typically designed to capture knowledge of the world corresponding to the mental model of an average human that is often implicit~\cite{Hayes1990TheNP,grice1975logic,Hespos2004ConceptualPT} and not directly learnable from text or images alone.

While the first large-scale commonsense resources such as Cyc~\cite{lenat1995cyc} utilized specialized internal representations (e.g. in CycL) to encode knowledge, more recent resources such as ConceptNet~\cite{liu2004conceptnet,speer2017conceptnet} and ATOMIC~\cite{hwang2021comet,sap2019atomic}, encode knowledge as free-text tuples of form \texttt{(head, relation, tail)} as part of a larger knowledge graph. The relative simplicity of the latter representation scheme has resulted in them being effective resources for integrating commonsense in neural language~\cite{yasunaga2022dragon,yasunaga-etal-2021-qa,young2018augmenting,wang-etal-2020-connecting} and vision models~\cite{karthik2022kg,zareian2020bridging,roy2022improving,Chen_Huang_Takamura_Nakayama_2021}, designing appropriate `reasoning' benchmarks~\cite{Gao2022Comfact,talmor-etal-2019-commonsenseqa,jin-etal-2022-cogkge} and being utilized to enhance `reasoning abilities' of large-scale language models~\cite{hwang2021comet,guan-etal-2020-knowledge,Bhargava2022CommonsenseKR}. 

\begin{figure}[t]
\includegraphics[width=1.01\columnwidth]{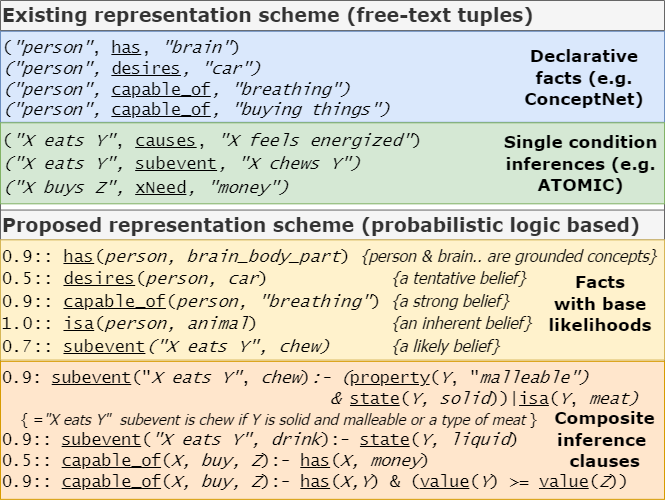}
\caption{Our proposed representation scheme. Facts represent base beliefs labelled with discrete certainty levels (e.g. `person desires car' is tentative while `person can breathe' is more certain and `person is an animal' is inherent). Composite clauses allow for more nuanced and dynamic inferences based on antecedents (e.g. sub-event of `X eats Y' has base tentative belief as `chew' which changes if knowledge of Y is available; see similarly for `X can buy Z'). Further, knowledge can be encoded flexibly using both free-text phrases (e.g. \textit{``X eats Y", ``breathing"}) and grounded concepts (e.g. \textit{person}, \textit{solid}, \textit{buy})} 
\label{fig:intro1}
\end{figure}

However, a prominent limitation of the free-text \texttt{(head, relation, tail)} representation scheme is that the knowledge it can express remains limited to declarative facts as in ConceptNet~\cite{speer2017conceptnet} (e.g. \texttt{(`human', `capable\_of', `buying things')}) or single-condition inferential knowledge as in ATOMIC (e.g. \texttt{(`X eats Y', `has\_subevent' `X chews Y')}). 
Such knowledge although useful at a surface level, only constitutes a portion of human commonsense which is often more nuanced and contextually-informed. For example, current CKBs cannot encode conceptual dependencies such as ``\texttt{a human can buy an object only if the money or assets they possess are greater than cost of the object}" or contextual dependencies such as ``\texttt{if Y is a liquid, then the subevent of X eats Y is much more likely to be X drinks Y (and much less likely to be X chews Y)}". Further, the tuple representation scheme does not encode likelihoods to differentiate certainties in knowledge. Consequently, as shown in fig. \ref{fig:intro1}, beliefs such as \texttt{(`human', `desires', `car')} or \texttt{(`X eats Y', `subevent' `X chews Y')}, that should be more tentative, are treated with the same likelihood to beliefs such as \texttt{(`human', `has', `brain')} or \texttt{(`X eats Y', `causes', `X gains energy')}, that should presumably be more certain. 

\textbf{Need for probabilistic-logic based representations of commonsense}.
One may look to address these limitations by simply representing conceptual or contextual antecedents in a single-text phrase (e.g. \texttt{(`\{X eats Y\} and \{Y is liquid\}', `subevent', `X drinks Y')}) and by adding a likelihood variable (e.g. \texttt{(`human', `capable\_of', `buying things', 0.6)}). However, this still does not effectively capture the composite nature of such knowledge wherein the inference and its likelihood is derived dynamically from a logical computation of relevant antecedents. 
Hence, for a more comprehensive representation scheme 
, we propose the utilization of probabilistic logic programming~\cite{de2015probabilistic} wherein, as illustrated in fig. \ref{fig:intro1}, base or context-independent conceptual beliefs can be represented as probabilistic facts (e.g. \texttt{0.95::has(person,brain)}; \texttt{0.6::desires(person,car)}) and composite or contextually-derived beliefs can be represented as probabilistic clauses (e.g. \texttt{0.95::capable\_of(X,buy,Z):-has(X,Y) \& value(Y)>=value(Z)} which states that \textit{``X can buy Z if X has Y and value of Y is more than equal Z's value"}). 

\textbf{Identifying salient relations and structuring knowledge with a hierarchical conceptual ontology}. Another limitation of prominent CKGs is that they utilize a small list of relations shared across all concepts (e.g. ConceptNet (v5.7)~\cite{speer2017conceptnet} has 36 relations and ATOMIC-2020~\cite{hwang2021comet} has 11). As noted by previous work~\cite{hwang2021comet,Bhargava2022CommonsenseKR}, this effectively limits the variety of commonsense captured by these CKGs resulting in them missing out on nuanced concept knowledge such as affordances, social and situational properties that form a vital portion of human commonsense and are also more difficult to capture through large-scale pretraining. Further, these CKGs lack a coherent knowledge organization structure and collect knowledge for a concept by only looking at the head phrase (of a tuple) in isolation without considering previously collected knowledge for related or parent concepts. This results in redundancies and sparsity across the larger knowledge graph.

\begin{figure}[t]

\includegraphics[width=0.99\columnwidth]{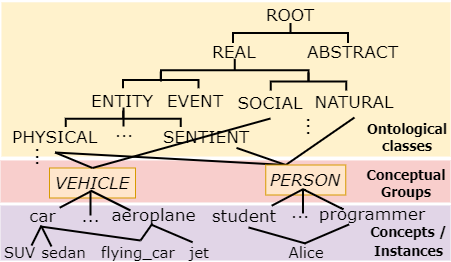}
\caption{Proposed hierarchical organization structure (condensed for visualization) to arrange concepts and identify salient concept-relevant relations and beliefs (refer table \ref{table:intro:ont_table}). Upper-level nodes reflect broad ontological classes from which more specific conceptual groups like `Vehicle' and `Person' are derived. Concepts (and their instances, e.g. Alice) inherit relations and beliefs from applicable parent conceptual groups (e.g. `flying\_car' inherits from both `car' and `aeroplane') which promotes re-use of knowledge when adding new concepts.} 
\label{fig:intro:ontology}
\end{figure}

To address these limitations, we propose the incorporation of a multi-level hierarchical conceptual ontology inspired by past cognitive models of human concept categorization~\cite{collins1969retrieval,rosch1973natural}. As shown in fig. \ref{fig:intro:ontology}, such a hierarchical ontology proceeds from broad ontological classes (e.g. REAL, MANMADE, PHYSICAL, etc) to relatively more specific ``conceptual groups"~\cite{dahlgren1986kind} (e.g. VEHICLE, PERSON, etc) from which finally actual concepts are derived (e.g. Vehicle -- \{Car, Aeroplane, ..\}) which can be made progressively more specific (e.g. Car -- \{SUV, Sedan, ..\}). As shown in table \ref{table:intro:ont_table}, to each concept node are attached relations and base beliefs that are inherited by their children nodes (unless over-ridden). This enables relations to be identified specific to conceptual-nodes (rather than globally for the entire knowledge graph) and to be progressively inherited downwards (e.g. `Vehicle' has specific relations \{mileage, top\_speed, travel\_area\} and also inherits relations from parents such as `Physical': \{size, location, etc\} and `Manmade':\{usage, construction, etc\}). Consequently, two lower-level concepts can have different applicable relations depending on their parents (e.g. `Car' shares `Physical' relations with `Programmer' but not `Vehicle' relations).



Similarly, conceptual beliefs (facts/clauses) can also be encoded at different conceptual levels and inherited downwards unless over-ridden (e.g. ``(X, can, think)" and ``(X, has, feelings)" is inherited by `Programmer' node from its parent `Sentient' and more specific beliefs such as ``(programmer, uses, terminal)" are specified at the `Programmer' node).
This effectively promotes re-use of knowledge when adding new lower-level concepts (similar to human few-shot concept learning) and helps better identify concept-specific beliefs with lesser resultant redundancy across the knowledge-graph.

\begin{table}[t]
\small
\setlength\tabcolsep{1.8pt}
\begin{tabular}{|l|c|c|c|} 
\hline
\rowcolor{Gray}
L & Node & Relations & Beliefs\\
\hline
0 & Root & \makecell{isa; can;\\ related\_to} & \makecell{isa(X,Z):-\\isa(X,Y),isa(X,Z)} \\
\hline
1 & Real & \makecell{comprises; \\created\_through ..}  & \makecell{exists\_in(X,world)}  \\
\hline
2 & Event & \makecell{causes; duration; \\subevent ..}  & \makecell{causes(X,change)}  \\
\hline
2 & Manmade & \makecell{used\_for; used\_by; \\ construction .. } & \makecell{created\_through(X,\\manufacturing); \\ 0.9: used\_by(human)} \\
\hline
3 & Physical & \makecell{location; material; \\ phy\_state; velocity ..}
& \makecell{made\_of(X,matter); \\in(X,Z):-in(X,Y),in(Y,Z)} \\
\hline
3 & Sentient & \makecell{desires; believes; \\ ment\_state; lang..}
& \makecell{can(X,think)\\has(X,feeling)} \\
\hline
6 & Vehicle & \makecell{mileage; top\_speed \\ travel\_area..}
& \makecell{used\_for(X,travel)\\requires(X,energy)} \\
\hline
7 & Programmer & \makecell{prog\_lang; \\ stack\_type..}
& \makecell{uses(X,terminal)\\can(X,program)} \\
\hline 
\end{tabular}\caption{Example ontology wherein relations and beliefs are identified specific to concept nodes and inherited downwards by children nodes (refer fig. \ref{fig:intro:ontology}). This allows for identification of concept-salient predicates (e.g. Vehicle has relations such as mileage and travel\_area besides inheriting `used\_for', etc from Manmade and `shape', etc. from Physical). It also promotes re-use of knowledge downwards (unless overridden) and reduces redundancies (e.g. Programmer inherits beliefs from Sentient, Physical, etc. allowing more specific beliefs to be identified for it). L refers to node level; more elaborate table in sec. \ref{sec:approach}.}\label{table:intro:ont_table}

\end{table}

\textbf{Probabilistic-logic based commonsense representation framework.} Based on the above two insights of utilizing a probabilistic-logic knowledge representation scheme and incorporating a hierarchical conceptual ontology, we propose a probabilistic-logic based commonsense representation framework. 
The framework is designed to model a wide variety of world knowledge and composite inferences including temporal verb-schemas, situational beliefs, physics laws and higher-order beliefs (as detailed in sec. \ref{sec:approach}). Its representation scheme is flexible such that knowledge can be represented through logical combinations of both grounded concepts and free-text phrases (as shown in fig. \ref{fig:intro1} for \textit{subevent(``X eats Y", chew)} and its antecedents -- e.g. ``malleable" is free-text while `solid' and `meat' are grounded concepts). In effect, our framework builds upon existing free-text phrase representation schemes and can be used to enrich existing CKGs with a further variety of knowledge and reasoning capabilities. In this work, we specifically extend the PrimeNet knowledge-base~\cite{primenet} with our framework, and demonstrate its usage as a grounded logic-based inference system more suitable for neuro-symbolic applications.



The rest of the paper is organized as follows -- (i) first, we review relevant work, (ii) second, we describe ProbNet in detail and how we construct a first version of it through both crowdsourcing and manual annotation, (iii) third, we illustrate its inference engine and how it can be applied for deeper and more interpretable passage-based semantic parsing and question answering, and (iv) finally, we discuss limitations of our current work and future directions. 

\vspace{-0.5em}
\section{Related work}
\label{sec:related_work}
\vspace{-0.2em}
\subsection{Commonsense knowledge resources and graphs}
Initial works in creation of machine-readable commonsense resources include the Cyc knowledge base~\cite{lenat1995cyc} and the Open Mind Commonsense Sense project~\cite{singh2002open} which evolved into ConceptNet\cite{liu2004conceptnet,speer2017conceptnet}. While Cyc attemped to hand-code commonsense knowledge in specialized `CycL' representations, ConceptNet obtained commonsense knowledge from the general public in text-phrase tuple forms through appropriate crowdsourcing mechanisms. 
Given its larger coverage and relatively simple representation scheme, ConceptNet has found greater utilization with recent neural (deep learning) approaches to improve `reasoning' abilities. Other prominent knowledge resources have also utilized relatively simpler representation schemes than Cyc, resulting in them being able to encode knowledge at a large-scale and be utilized with neural methods. Prominent examples include SenticNet~\cite{cambria2022senticnet} for modelling knowledge relevant to sentiment analysis tasks, WebChild~\cite{tandon2014webchild} and ASER knowledge graph~\cite{zhang2020aser}
that extract information through web crawling, ATOMIC~\cite{sap2019atomic,hwang2021comet} which models `if-then' inferential knowledge for event situations and CSKG~\cite{ilievski2021cskg} that encodes knowledge across different such resources to obtain a unified knowledge graph. For a more comprehensive survey, we refer the reader to~\cite{storks2019recent}.

In relation to these works, our work focuses on how knowledge can be better represented by utilizing probabilistic logic to model composite and dynamic inferences, and incorporating a hierarchical ontology to structure beliefs and identify salient relations. We believe this can be a complementary direction to the development of further commonsense resources or updates in existing ones, particularly given the growing developments in neuro-symbolic learning methods~\cite{Sen2022NeuroSymbolicIL,glanois2022neuro,moghimifar-etal-2021-neural} and their utilization for reasoning tasks~\cite{Winters2022DeepStochLogNS,manhaeve2018deepproblog,mao2019neuro}.






\subsection{Commonsense datasets and neural knowledge graphs}
An alternative direction to developing commonsense knowledge resources in form of knowledge-graphs or -bases has been to curate datasets for specific types of commonsense reasoning. These datasets can then be used to train neural or large-scale pretrained models to better capture the targeted type of commonsense, besides serving as benchmarks for the same.  Some prominent works in this direction include Piqa~\cite{bisk2020piqa} for physical commonsense, VCR for visual commonsense~\cite{zellers2019recognition}, CICERO~\cite{ghosal-etal-2022-cicero} for contextually-informed dialogue inferences, SocialQA~\cite{sap-etal-2019-social} for social situational commonsense and TIMEDIAL~\cite{Qin2021TIMEDIALTC} for temporal commonsense for dialogue inferences. Further, works~\cite{hwang2021comet} have also looked into how commonsense resources can be combined with large-scale pretrained language models (see~\cite{Bhargava2022CommonsenseKR} for an extensive survey) to be represented as `neural knowledge graphs' and queryable through prompting. In relation to the above directions, we believe our work can support development of newer datasets / benchmarks that evaluate composite logical processing (similar to how CommonsenseQA~\cite{talmor-etal-2019-commonsenseqa} utilizes ConceptNet) besides being utilized to potentially enable neural knowledge graphs to make contextually-dependent and logical inferences with varying likelihoods / certainties.



\subsection{Answer set programming frameworks}
Answer set programming (ASP)~\cite{lifschitz2019answer,Gelfond2014KnowledgeRR} is another prominent logic-based knowledge representation and reasoning approach. In ASP, given  knowledge (rules and facts), valid models (specifically `stable Herbrand models' that contain no variables) are generated from which solutions can be inferred. This is a bottom-up model generation approach in contrast to traditional Prolog-like logic programming wherein given a query, backward chaining is performed with matching rules/facts to prove (or disprove) the query. Recent works have shown how ASP can be utilized for natural language inference tasks~\cite{basu2021knowledge}. However, in contrast to probabilistic-logic frameworks such as ProbLog, prominent ASP frameworks do not allow for representing uncertainties in inference and lack neuro-symbolic integration, both of which motivate our choice to use probabilistic logic programming for our work. 







\section{Probabilistic-logic based commonsense representation framework}
\label{sec:approach}
We aim to design a knowledge representation and reasoning scheme that can effectively model inferential knowledge comprising multiple antecedents and represent beliefs with varying likelihoods. Additionally, our target representation scheme should allow for the knowledge to be utilized as both a free-text knowledge-graph and a grounded semantic inference system. 
Accordingly, as mentioned previously, we will utilize a probabilistic logic programming approach to model such knowledge and perform inferences. Specifically, we utilize the ProbLog language~\cite{Raedt2007ProbLogAP,de2015probabilistic} for our work given its well-developed framework~\cite{Fierens2014InferenceAL,dries2015problog2}, and its extensions for relational learning and neuro-symbolic processing~\cite{manhaeve2018deepproblog}, which are both relevant to future directions of our work on learning such knowledge.

\subsection{Probabilistic logic programming primer / syntax}
Probabilistic logic programming extends traditional logic programming~\cite{lloyd2012foundations} (as in Prolog) with probabilities attached to facts and clauses to model uncertainties. Table \ref{table:appendix:primer} (appendix) lists probabilistic logic programming terminology and ProbLog syntax relevant to our work's discussion. An example representation of the natural language belief \textit{`X can move with base likelihood 0.9 if X has a leg or a wheel and X is not in a stuck state'} in logical form is ``\texttt{0.9::can(X,move):- (has(X,leg); has(X,wheel)), not(state(X,stuck)).}" 
For a more comprehensive overview of probabilistic logic programming, we refer the reader to~\cite{de2015probabilistic}.

\subsection{Extending PrimeNet with a base conceptual ontology and typed predicates}
\begin{figure}[t]
\includegraphics[width=0.99\columnwidth]{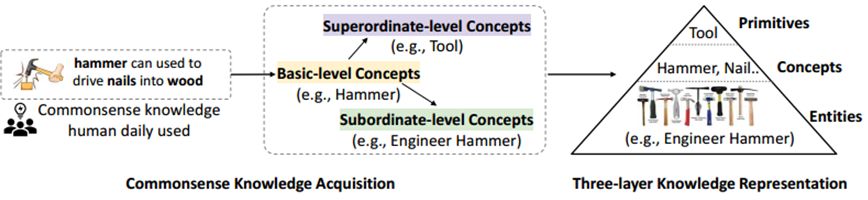}
\caption{Existing three-layer knowledge representation of PrimeNet.} 
\label{fig:approach:primenet}
\end{figure}

PrimeNet is a commonsense knowledge base organised hierarchically around a basic level of concepts, in terms of which humans tend to think about the world.  It was constructed by organising these basic level concepts under a superordinate level of much fewer primitive classes (hence the name PrimeNet), and over a subordinate level of many more entity classes which are specialisations of the basic level concepts. PrimeNet is psychologically inspired from observations that humans rely on a much smaller set of concepts to function in the world compared to the millions of lexical concepts that exist, and hypothesises that commonsense reasoning could similarly depend on a concise core of primitive concepts as shown in fig. \ref{fig:approach:primenet}. 

Our commonsense knowledge representation and reasoning scheme extends PrimeNet as follows:
\begin{enumerate}
    \item We ground the primitive level classes in PrimeNet to a base ontology adapted from Dahlgren~\cite{dahlgren1986kind} as previously introduced in fig.\ref{fig:intro:ontology}. We do this by equating conceptual primitives in PrimeNet to Kind Types in the base ontology.  
    \item We introduce type-specific predicates (relations and attributes) to each concept node that are inherited downwards by subordinate concepts.
    \item We associate probabilities with each conceptual belief (facts/clauses) to reflect the likelihood that people would hold such a belief.
\end{enumerate}

Our ontology's hierarchy was designed specifically to comprise a lesser number of nodes and levels in comparison to the more extensive WordNet hierarchy~\cite{miller1995wordnet}. This was done as our aim was to utilize the hierarchy to identify concept-salient predicates and encode beliefs at different levels in a manageable manner. A subset of the ontology with salient predicates is shown in table \ref{table:approach:ontology_and_predicates}. With such a structure, new predicates, beliefs and concepts can be added incrementally at lower-levels to gradually increase the variety of knowledge expressible. For events, as we detail later, in addition to conceptual predicates such as `causes', `enables', etc., we utilize semantic roles drawn from VerbNet~\cite{schuler2005verbnet} to represent particular instances of events and any beliefs pertaining to entities.


\vspace*{-5em}
\begin{center}

\small{
\setlength\tabcolsep{2pt}
\begin{tabular}{|c|c|c|l|l|} 
\hline
\rowcolor{Gray}
L & Nd & Parents & Predicates & Examples\\
\hline
0 & \rotatebox[origin=c]{90}{Root} & - & \makecell[l]{isa \\ can} & \makecell[l]{isa(car,vehicle)\\can(article, ``transmit ideas")} \\
\hline
1 & \rotatebox[origin=c]{90}{Real} & Root & \makecell[l]{comprises \\ created\_by} & \makecell[l]{comprises(book,sentences)\\created\_by(car,construction)} \\
\hline
2 & \rotatebox[origin=c]{90}{Numeric} & \makecell[l]{Abstract\\-$>$Root} & \makecell[l]{value \\ more\_than} & \makecell[l]{value(temp,30,celsius)\\ more\_than(wtr\_bp,room\_tmp)} \\
\hline
3 & \rotatebox[origin=c]{90}{Idea} & \makecell[l]{Proposition\\-$>$Abstract} & \makecell[l]{content \\ author} & \makecell[l]{content(theory,axioms)\\ author(motion\_law,``Newton")} \\
\hline
2 & \rotatebox[origin=c]{90}{Event}  & \makecell[l]{Real\\-$>$Root} & \makecell[l]{causes \\ duration \\ theme \\ subevent \\ purpose} & \makecell[l]{causes(heat\_liq, evaporation)\\ duration(make\_home,`months')\\ theme(eat,food) \\ subevent(buy, negotiate) \\ purpose(``fix item",``use item")} \\
\hline
4 & \rotatebox[origin=c]{90}{Cycle}  & \makecell[l]{Process\\-$>$Event..} & \makecell[l]{trigger\\ sequence} & \makecell[l]{trigger(brayton\_cycle, spark), \\seq.(rain\_cy,[evap,cond,precip])}\\
\hline 
2 & \rotatebox[origin=c]{90}{Manmade}  & \makecell[l]{Real\\-$>$Root} & \makecell[l]{used\_for\\ used\_by} & \makecell[l]{used\_for(car,travel) \\used\_by(wrench,mechanic)}\\
\hline
3 & \rotatebox[origin=c]{90}{Physical}  & \makecell[l]{Entity\\-$>$Real..} & \makecell[l]{location\\ phy\_state \\ has\_part\\has\_aspect\\temperature } & \makecell[l]{location(car,garage) \\phy\_state(water, liquid)) \\has\_part(car, wheels, 4) \\has\_aspect(cube, surface, 6)\\temp.(moon, 120, celsius)}\\
\hline
3 & \rotatebox[origin=c]{90}{Sentient}  & \makecell[l]{Entity\\-$>$Real..} & \makecell[l]{mental\_state\\ desires \\ has\_trait} & \makecell[l]{mental\_state(person,happy) \\desires(victim,justice) \\ trait(politician,`good speaker')}\\
\hline
4 & \rotatebox[origin=c]{90}{Living}  & \makecell[l]{Physical\\-$>$Entity..} & \makecell[l]{age\\lifespan} & \makecell[l]{age(cat1,12) \\lifespan(dog,10,14)}\\
\hline
5 & \rotatebox[origin=c]{90}{Animal}  & \makecell[l]{Living,\\-$>$Physical..} & \makecell[l]{gender\\behavior} & \makecell[l]{gender(hen,female) \\behavior(lion,``territorial")}\\
\hline
5 & \rotatebox[origin=c]{90}{Fluid}  & \makecell[l]{Non-living,\\-$>$Physical..} & \makecell[l]{viscosity\\boil\_point} & \makecell[l]{viscosity(water, 0.01, poise) \\boil\_pt.(liq\_oxygen,-194, cels.)}\\
\hline
5 & \rotatebox[origin=c]{90}{Device}  & \makecell[l]{Non-living; \\ManMd.-$>$Real..} & \makecell[l]{energy\_type \\power\_used} & \makecell[l]{energy\_type(fridge,electric) \\power\_used(bulb,50, watts)}\\
\hline
6 & \rotatebox[origin=c]{90}{Vehicle}  & \makecell[l]{Device,\\-$>$Manmade..} & \makecell[l]{travel\_area \\ mileage} & \makecell[l]{travel\_area(plane,``continental") \\mileage(sport\_bike, 4, km$/$ltr)}\\
\hline


\end{tabular}\label{table:approach:ontology_and_predicates}
\captionof{table}{Subset of ontology  (L and Nd refer to level \& node) with examples of salient predicates (4th col.) for conceptual groups.}
}
\end{center}


\subsection{Knowledge representation and reasoning scheme}
To illustrate our knowledge representation and reasoning scheme, we make use of the format as in table \ref{table:approach:example_inference} where `Knowledge' refers to encoded facts and beliefs, `Example queries' indicate inference or retrieval-time queries and `Inferences' indicate their corresponding results. Note, that knowledge indicated in further examples are to illustrate the representation scheme capabilities with familiar concepts / examples and not all are necessarily encoded in the first version of ProbNet's knowledge-base version (which is constructed for aerospace concepts as detailed in sec. \ref{sec:applications}).

\begin{table}[h]
\small
\setlength\tabcolsep{3pt}
\begin{tabular}{|wl{0.05\columnwidth}|wl{0.9\columnwidth}|}
\rowcolor{Gray}
\hline
\multicolumn{2}{|l|}{Knowledge} \\ 
\hline
F1 & 0.6::can(animal,move).\\
F2 & 0.8::has(car,wheel). \\
C1 & 0.9::can(X,move):-has(X,leg); has(X,wheel). \\
\rowcolor{Gray}
\hline 
\multicolumn{2}{|l|}{Example queries} \\ 
\hline
Q1 & can(car, move)?\\ 
Q2 & can(X, move)? \\
\rowcolor{Gray}
\hline 
\multicolumn{2}{|l|}{Inferences} \\ 
\hline
I1 & 0.72: can(car, move). \\
I2 & 0.6: can(X=animal, move), 0.72: can(X=car, move).\\
\hline

\end{tabular}\caption{Example format used to illustrate representation scheme and inferences (F1 and F2 are distinct facts, C1 is a clause, Q1 and Q2 are distinct queries while I1 and I2 are their corresponding inferences).}\label{table:approach:example_inference}
\end{table}

\textbf{General representation scheme.}
We represent \textbf{declarative facts} in the following format: \par

\texttt{P::predicate([fact\_id], [source\_id], time\_point, pred\_arg1, pred\_arg2, ..).}\par 

Here \texttt{P} refers to probability of fact, \texttt{[fact\_id]} refers to a unique fact identifier, \texttt{[source\_id]} refers to source of fact, \texttt{time\_point} indicates the time at which a fact is valid and the arguments thereafter are the original predicate specific arguments. While \texttt{P} of a fact can be a continuous number between 0 to 1, in this work we utilize four discrete `certainty' levels with corresponding \texttt{P}. These are -- (i) \textit{tentative} (\texttt{P=0.5}), (ii) \textit{likely} (\texttt{P=0.7}), (iii) \textit{strongly likely} (\texttt{P=0.9}), and (iv) \textit{inherently true} (\texttt{P=1.0}). 

The fact and source identifiers are both used for interpretability when multi-hop inferences are performed with different knowledge sources (detailed later). While one could achieve inference trace through just the fact\_id, we maintain a distinct source argument to enable source-dependent logic (e.g. some knowledge sources might be known to be unreliable and thereby be given low likelihood). Further, the trace retrieval can also be used to differentiate cases wherein a query is evaluated False (or likelihood=0) due to absence of any matching knowledge (in which case trace is empty) or due to a rule/fact activation (in which case trace is not empty). This effectively \textbf{relaxes the closed-world processing assumption}.

The time point indicator is introduced for temporal and event logic (detailed later), and the default time point is \texttt{t\_g} denoting general time (indicating a fact is assumed to be generally valid irrespective of time unless over-ridden for a particular time point). 

We represent \textbf{inference clauses} with N antecedents in the following format: \par
\texttt{P::head\_predicate(Fh,Sh,Th,args):-}\par\texttt{anc1\_pred(F1,S1,T1,a1\_args..),..} \par \texttt{ancN\_pred(FN,SN,TN,aN\_args..),} \par\texttt{union\_ops(Fh,C\_id,F1..FN),} \par\texttt{union\_ops(Sh,S1..SN).}

The latter two `union\_ops' refer to list operations (detailed in \ref{table:approach:base_scheme} through Prolog operators `append' and `sort') to derive Fh and Sh through union of F1 to FN and S1 to SN respectively.  C\_id refers to the unique clause id which is added for interpretability trace.


%
\begin{table}[h]
\small
\setlength\tabcolsep{3pt}
\begin{tabular}{|wl{0.05\columnwidth}|wl{0.9\columnwidth}|}
\rowcolor{Gray}
\hline
\multicolumn{2}{|l|}{Knowledge} \\ 
\hline
F1 & 1::isa([f1],[`wnt'], t\_g, person, organism). \textit{\%facts}\\
F2 & 1::isa([f2],[`kb'], t\_g, programmer, person). \\
F3 & 0::isa([f3],[`kb'], t\_g, person, car). \\
\hline
C1 & \makecell[l]{\textit{\%Example clause with trace}\\1::isa(F,S,T,X,Z):- isa(F1,S1,T,X,Y), isa(F2,S2,T,Y,Z),\\ \hspace{2.3cm} append([S1,S2], S3), sort(S3, S), \\
\hspace{2.3cm} append([F1, F2], F3), sort([\textbf{c1}$|$F3],F).} \\
\rowcolor{Gray}
\hline 
\multicolumn{2}{|l|}{Example queries} \\ 
\hline
Q1 & isa(F,S,T, person, organism)?\\ 
Q2 & isa(F,S,T, person, programmer)?\\
Q3 & isa(F,S,T, person, car)?\\
Q4 & isa(F,S,T, programmer, Y)? \\
\rowcolor{Gray}
\hline 
\multicolumn{2}{|l|}{Inferences} \\ 
\hline
I1 & 1: isa(F=[f1], S=['wnt'], T=t\_g, person, organism). \\
I2 & 0: isa(F=?,S=?,T=?, person, programmer) \textbf{\textit{\{unknown\}}}\\
I3 & 0: isa(F=[f3],S=['kb'],T=t\_g, person, car) \textbf{\textit{\{known\}}} \\
\hline
I4 & \makecell[l]{1: isa(F=[f2], S=['kb'], T=t\_g, programmer, Y=person) \\
               1: isa(F=[f2,c1,f1],S=['kb','wnet'],T=t\_g, programmer,\\ 
               \hspace{0.8cm}Y=organism)} \\
\hline

\end{tabular}\caption{\textbf{Base representation scheme with inference traces}. As shown for inference I4, the 1st derivation (Y=person) is direct fact lookup (F2), while the 2nd derivation (Y=organism) applies clause C1 on F2 and then finds F1. Further, \textbf{traces relax closed-world assumption} as shown for I2 where isa(person,programmer) is indicated unknown, cf. I3 where isa(person,car) is known from F3 to be false.}\label{table:approach:base_scheme}
\end{table}

An example utilization of the above scheme is shown in table \ref{table:approach:base_scheme}. Two basic facts of isa(person, organism) and isa(programmer, person) are encoded with the first obtained from source `wnet' and second from source `kb'. An inference rule for inheritance is represented as shown in C1 (the basic rule being isa(X,Z):-isa(X,Y),isa(Y,Z)), and augmented with list operations for trace encoding. The first query Q1 is solved through direct retrieval of F1, the second query Q2 is indicated to be unknown as no matching facts were found and the third query Q3 is indicated to be False based on fact F3. Finally, Q4 which asks what all things (variable `Y') a programmer is, is resolved through retrieval of F2 and then application of C1 and retrieval of F1. 

For the remainder of our paper, we omit indication of variables F, S and T in facts or clauses unless relevant.



\textbf{Hierarchical knowledge inheritance.}
As shown in table \ref{table:approach:hierarchy}, beliefs can be attached at different levels (e.g. an `animal' \textit{can} do a `motor action' is represented in F1, while `bird' \textit{can} `fly' is represented in F7). To inherit beliefs (or any predicate in general), we utilize the inheritance rules listed in C1 and C2. While C1 specifies that \textit{``X is Z if X is Y and Y is Z"}, C2 specifies that \textit{``if X is Z and Z can do Y, then X can do Y \textbf{provided that} no knowledge that X cannot do Y is mentioned".} As shown in Q1, the belief that a `sparrow' can do a motion activity is inferred through inheritance of properties from `sparrow's' parent node `animal' through application of C2 and C1. Similarly as shown in Q2, `duck' inherits beliefs that it can fly from parent `bird' and in Q6, it has its own concept-level belief that `duck' can `swim' (which is unknown for a `sparrow' as shown in Q7). 
While shown for the predicate \textit{`can'}, such inheritance can be applied with all existing predicates through the same rule-type as in C2.

To specify that a belief is not true, the `not' prefix is assigned to the existing predicate. As shown for F13, it is used to specify that a \textit{`penguin  cannot fly'}. Similarly, in F15, for a hypothetical `new\_bird', the inherited belief that `X can fly' is made tentative by adding the `not' prefix. One may attempt to override knowledge by changing C2's last antecedent to be simply \textit{not(can(X,Y))}; however this will lead to a cyclic loop not permitted in ProbLog. Similarly, simply setting the likelihood to 0 as done in F14 to specify \textit{`penguin2' cannot fly} will also not work since in cases wherein more than one trace exists (in this case C2 and F14), the belief is inferred through noisy-or application: $P(f) = 1 - P(f=False)$ where $P(f=False) = \prod_{k}{(1-P(trace\_k))}$. As a result, for Q4 the belief that \textit{``penguin2 can fly"} remains true with a probability of 0.9 (computed as 1- ((1-0)*(1-0.9)). In contrast, for Q3 it is 0 as the inheritance is blocked, while for Q5, the belief \textit{``new\_bird can fly"} has resultant probability of 0.45 (computed as 0.9*0.5).\\


\begin{table}[t]
\small
\setlength\tabcolsep{3pt}
\begin{tabular}{|wl{0.07\columnwidth}|wl{0.9\columnwidth}|}
\rowcolor{Gray}
\hline
\multicolumn{2}{|l|}{Knowledge} \\ 
\hline
F1 & 0.8::can(animal, motion\_activity).\\ 
F2 & isa(bird, animal).\\
F3-4 & isa(fly, motion\_activity). isa(swim, motion\_activity).\\
F7 & 0.9::can(bird, fly).\\
F8-10 & isa(sparrow, bird). isa(duck, bird), isa(penguin, bird).\\
F11 & isa(penguin2, bird).  \\
F12 & isa(new\_bird, bird).\\
F13 & not\_can(penguin, fly). \textit{\%Overrides inherited fact}\\
F14 & 0::can(penguin2, fly). \textit{\%Will not override}\\
F15 & 0.5::not\_can(new\_bird, fly). \textit{\%Override to tentative}\\
F16 & 0.9:: can(duck, swim). \\
\hline
C1 & isa(X,Z):- isa(X,Y),isa(Y,Z). \textit{\%basic inheritance clause}\\

C2 & \makecell[l]{\textit{\%Below is property inheritance (unless exception)} \\can(X,Y):-isa(X,Z),can(Z,Y), not(not\_can(X,Y)).}\\
\hline

\rowcolor{Gray}
\hline 
\multicolumn{2}{|l|}{Example queries} \\ 
\hline
Q1 & can(sparrow, motion\_activity)?\\ 
Q2,3 & can(duck, fly)? can(penguin, fly)?\\
Q4,5 & can(penguin2, fly)? can(new\_bird, fly)? \\
Q6,7 & can(duck, swim)? can(sparrow, swim)? \\
\rowcolor{Gray}
\hline 
\multicolumn{2}{|l|}{Inferences} \\ 
\hline
I1 & 0.8 can(sparrow, motion\_activity). \{\textit{from C2,F8,C1,..F1}\}\\
I2,3 & 0.9 can(duck, fly). 0 can(penguin,fly).\\
I4,5 & 0.9 can(penguin2, fly). 0.45 can(new\_bird, fly). \\
I6 & 0.9 can(duck, swim). \\
I7 & 0 can(sparrow,swim). \textit{\{unknown\}}\\
\hline

\end{tabular}\caption{\textbf{Hierarchical knowledge inheritance}. Clauses C1 and C2 enable beliefs to be inherited downwards such as for I1 where two-hop computation is performed. F13 overrides `can fly' belief for penguin by using not\_can predicate (F14 attempts to override by setting likelihood of fact to be zero but will not work due to noisy-or inference). F15 makes inherited belief `can fly' for `new\_bird' more tentative.}\label{table:approach:hierarchy}
\end{table}

\textbf{Grounded concepts and free-text phrase representations.} In our representation scheme, we distinguish between free-text phrases (denoted in ``inverted commas") from grounded concepts.  Whereas free-text phrases are merely lexical labels, grounded concepts are concepts in our knowledge graph with associated semantic properties that can be used for inference.  We illustrate this distinction by first describing how basic lookup of a concept is performed by a free-text name (lexicon lookup). As shown in table \ref{table:approach:free_text_lookup}, F1 refers to a grounded concept \textit{bowl}, which is specified as a type of container. F2 and F3 state that the concept \textit{bowl} can be referred to by multiple free-text names ``bowl" and ``basin" (the latter with lesser probability). However, the name ``bowl" can also refer to other applicable concepts, e.g. the action ``roll" or ``bowling" (see F7).

Querying concept beliefs can then be performed in two ways -- (i) specify the actual concept (e.g. \textit{bowl}) as in Q2, or (ii) specify the free-text phrase to retrieve applicable concepts first (as in Q1, Q3, Q4) and then decide which concept to utilize (effectively performing word-sense disambiguation as in Q3 and Q4 where conditions on applicable concepts are provided). This is in contrast to CKRs wherein concepts and free-text phrases are treated as the same, which can potentially lead to  duplications and inter-mixing of knowledge (e.g. beliefs pertaining to the event ``bowl" or ``roll" might be combined with those of the entity ``bowl"). 

\begin{table}[h]
\small
\setlength\tabcolsep{2.5pt}
\begin{tabular}{|wl{0.04\columnwidth}|wl{0.95\columnwidth}|}
\rowcolor{Gray}
\hline
\multicolumn{2}{|l|}{Knowledge} \\ 
\hline
F1 & isa(bowl, container).\\ 
F2 & has\_name(bowl, ``bowl").\\
F3 & 0.8::has\_name(bowl, ``basin").\\
F4 & 0.6::used\_for(bowl, ``eating soup"). \textit{\%symbol $+$ phrase fact}\\

F5 & isa(roll\_action, event).\\
F6 & has\_name(roll\_action, ``roll').\\
F7 & 0.6::has\_name(roll\_action, ``bowl'). \textit{\%e.g. bowl a ball} \\

F8 & can\_be(newobj, close\_state). \textit{\%i.e. can be closed}\\
F9 & has(newobj, ``solid enclosure"). \\

\hline
C1 & 0.6::used\_for(X,phy\_storage):- isa(X, container). \\
C2 & \makecell[l]{0.7::can(X,``keep things"):- \textit{\%phrase $+$ symbol clause}\\ can\_be(X, close\_state), has(X, ``solid enclosure").} \\
C3 & 1::can(X,``keep things"):-used\_for(X,phy\_storage).\\

\hline

\rowcolor{Gray}
\hline 
\multicolumn{2}{|l|}{Example queries} \\ 
\hline
Q1 & has\_name(X,``bowl")? \%\textit{What concepts have name ``bowl"}\\  
Q2 & has\_name(bowl, Y)? \%\textit{What names does concept \underline{bowl} have}\\
Q3 & has\_name(Q,``bowl"),isa(Q,event)? \textit{\%event named `bowl'?} \\ 
Q4 & has\_name(Q,``bowl"), used\_for(Q,Y)?  \textit{\%exec. in sequence}\\ 
Q5 & can(X, ``keep things")?  \%\textit{Free-text / phrase query}\\

\rowcolor{Gray}
\hline 
\multicolumn{2}{|l|}{Inferences} \\ 
\hline
I1 & \makecell[l]{1.0 has\_name(bowl, ``bowl"), \\ 0.6 has\_name(roll\_action, ``bowl").}\\
I2 & 1.0 has\_name(bowl, ``bowl"), 0.8 has\_name(bowl, ``basin") \\
I3 & 0.6 has\_name(Q=roll\_action,``bowl"), isa(roll\_action,event) \\
I4 & \makecell[l]{1 has\_name(Q=bowl, ``bowl"), \\0.6 used\_for(bowl,phy\_storage), used\_for(..,``eating soup")} \\
I5 & \makecell[l]{0.7 can(X=newobj, ``keep things"), \\0.6 can(container, ``keep things"), can(bowl, ``keep things")} \\
\hline

\end{tabular}\caption{\textbf{Utilization of grounded concepts and free-text phrases for knowledge representation and lexicon lookup}. A concept can have multiple free-text names (e.g. in F2, F3 concept \underline{bowl} has names ``bowl" and ``basin") and conversely free-text names can refer to multiple concepts (e.g. free-text ``bowl" refers to concepts \underline{bowl(obj)} (F2) and \underline{roll(event)} (F7)). Further, \textbf{clauses \& facts can be queried or represented using both constants \& free-text} (e.g. Q5 queries X that can ``keep things"; similarly F4,F9 \& C2,C3 for such facts \& clauses).}\label{table:approach:free_text_lookup}
\end{table}

More generally, beliefs for concepts can be encoded by using both free-text phrases and grounded concepts. As shown in F4 and F9, free-text phrases can be utilized when a grounded representation may not be yet feasible (e.g. attributes such as ``eating soup" or ``solid enclosure"). In C2 knowledge that X can be used to ``keep things" has both a grounded antecedent (X can be \textit{close\_state}) and  a free-text antecedent (X has a ``solid enclosure"). Similarly C3 captures a free-text phrase implication with a grounded antecedent (X can ``keep things" if X used for \textit{phy\_storage}). As a result, for Q5 which performs a free-text query (\textit{what objects can ``keep things"}), the concept \textit{newobj} is found (due to C2) in addition to \textit{container} and \textit{bowl} (due to C3, C1).


\begin{table}[h]
\small
\setlength\tabcolsep{2pt} 
\begin{tabular}{|wl{0.05\columnwidth}|wl{0.96\columnwidth}|}
\rowcolor{Gray}
\hline
\multicolumn{2}{|l|}{Knowledge} \\ 
\hline
F1 & 1::isa(eat, consume\_activity). \\
F2 & 0.9::theme(eat, food).\% base beliefs\\
F3 & 0.9::agent(eat, animal).\\
F4 & 0.5::subevent(eat, ``salivate").\\
\hline
C1 & \makecell[l]{0.9:: subevent(E1,chew):- isa(E1,eat), theme(E1,X), 
\\    (phy\_state(X,solid),property(X,``malleable")).}\\
C2 & \makecell[l]{0.9:: subevent(E1,``crunch"):- isa(E1,eat), theme(E1,X), 
\\    property(X,``crispy").}\\
C3 & \makecell[l]{0.9:: subevent(E1, drink):- isa(E1, eat), theme(E1,X), 
\\    phy\_state(X,liquid).}\\
C4 & \makecell[l]{0.9:: instrument(E1,cutlery):- isa(E1,eat), agent(E1,X), 
\\    isa(X,human), (loc(E1,fine\_dining); attrb(E1,``formal")).}\\
C5 & \makecell[l]{0.9:: instrument(E1,mouth):- isa(E1,eat), agent(E1,X), 
\\    (isa(X,cat); isa(X,dog)).}\\

%

\hline

\rowcolor{Gray}
\hline 
\multicolumn{2}{|l|}{Example queries (with background knowledge)} \\ 
\hline
Q1 & isa(eat\_ins, eat). \textcolor{orange}{agent(eat\_ins, A)? subevent(eat\_ins, E)? }\\
Q2 & \makecell[l]{isa(eat\_i, eat), theme(eat\_i, nachos).\\property(nachos, ``crispy"). \textcolor{orange}{subevent(eat\_i, E)?}} \\
Q3 & \makecell[l]{isa(bob, human). isa(eat\_1, eat). agent(eat\_1, bob).\\ isa(micheli, fine\_dining). location(eat\_1, micheli). \\ \textcolor{orange}{instrument(eat\_1,I?)?}}\\
Q4 & \makecell[l]{isa(husky, dog). isa(eat\_2, eat). agent(eat\_2, husky).\\ isa(micheli, fine\_dining). location(eat\_2, micheli). \\\textcolor{orange}{instrument(eat\_2,I?)?}}\\
\rowcolor{Gray}
\hline 
\multicolumn{2}{|l|}{Inferences} \\ 
\hline
I1 & 0.9 agent(eat\_ins, A=animal). 0.5 subevent(..E=``salivate").\\
I2 & 0.9 subevent(eat\_i, E=``crunch"). 0.5 subevent(..``salivate").\\
I3 & 0.9 instrument(eat\_1, I=cutlery). \\
I4 & 0.9 instrument(eat\_1, I=mouth). \\
\hline

\end{tabular}\caption{\textbf{Event and situational inferences using semantic role knowledge.} Facts F2-F4 represent base beliefs for event `eat' (e.g. F2 the theme or thing eaten is probably a type of \textit{food}). C1-C3 represent clauses indicating different possible subevents for `eat' depending on knowledge of item being eaten (e.g. C3 states that if item X being eaten is a liquid, then sub-event is more likely to be \textit{drink}). Similarly, C4 \& C5 state the instrument used to eat differs depending on agent or loc. of event (e.g. instrument is likely mouth if agent= dog or cat).}\label{table:approach:event_roles}
\end{table}


\textbf{\\Event representations and situational inferences.}
To represent event or situational knowledge, we utilize semantic-roles~\cite{gruber1965studies,fillmore1967case} (such as `theme', `agent', `instrument', etc.) that we identify for broad event categories in the ontology (utilizing VerbNet~\cite{schuler2005verbnet} wherever possible). An example is illustrated in table \ref{table:approach:event_roles} for the event \textit{`eat'} which is a type of \textit{`consume\_activity'} (F1). F2 and F3 represent base context-independent beliefs, more specifically, selectional preferences~\cite{resnik1997selectional} for the event, e.g. that the \textit{theme} of \textit{`eat'} is a type of \textit{`food'} (F2), and the \textit{agent} is a type of \textit{`animal'} (F3). F4 specifies a tentative belief that  subevent of eat is \textit{`salivate'}.

To capture contextual or situational dependencies, we make use of clauses that allow different inferences to be made on provided contextual/situational facts. For example, C1-C3 represent how the \textit{subevent} of \textit{`eat'} can differ depending on the properties of the \textit{`theme'} or the thing being eaten. C1 states that the subevent is most likely \textit{chew} if the theme X has physical\_state = solid and that it also has property= \textit{``malleable"}. C2 states that subevent is most likely \textit{``crunch"} if theme X has property=\textit{``crispy"} while C3 states subevent is most likely \textit{``drink"} if theme X is a type of liquid. 

Thus, depending on the background context information provided during querying, the inferences can vary. As shown for Q2, it is specified that the entity \textit{nachos} is being eaten and that \textit{nachos} are ``crispy". Consequently, the inferred subevent for this \textbf{instance} of \textit{eat} is probably \textit{``crunch"} in addition to \textit{``salivate"}. Similarly, C4-C5 capture inferences for type of instrument being used for \textit{``eat"} depending on agent or location of event. Q3 and Q4 illustrate two different instances of `eat' both happening at a \textit{fine\_dining} location `micheli'. In Q3, agent is \textit{bob} known to be a \textit{human} and hence inferred instrument is probably a type of \textit{cutlery}, while for Q4, the agent is a husky, and instrument is more likely to be simply \textit{mouth}. 




\begin{table}[t]
\small
\setlength\tabcolsep{2pt}
\begin{tabular}{|wl{0.05\columnwidth}|wl{0.97\columnwidth}|}
\rowcolor{Gray}
\hline
\multicolumn{2}{|l|}{Knowledge} \\ 
\hline
C1 & \makecell[l]{location(T3,X,D,in):- isa(t\_g,E,enter\_phy), theme(t\_g,E,X), 
               \\\hspace{2.1cm}destination(t\_g,E,D), t\_end(t\_g,E,T3).} \\
               
C2 & \makecell[l]{location(T1,X,S,at):-isa(t\_g,E,enter\_phy), theme(t\_g,E,X), \\\hspace{2.8cm}source(t\_g,E,S), t\_start(t\_g,E,T1).} \\  
               
C3 & \makecell[l]{location(T2,X,C,at):-isa(t\_g,E,enter\_phy), theme(t\_g,E,X), \\\hspace{2.8cm}channel(t\_g,E,C), t\_during(t\_g,E,T2).} \\     

C4 & \makecell[l]{location(T1,X,D,out):-isa(t\_g,E,enter\_phy), theme(t\_g,E,X), \\\hspace{2.8cm}destination(t\_g,E,D), t\_start(t\_g,E,T1).} \\     

C5 & \makecell[l]{more\_than(t\_g,Tm2,Tm1):-isa(t\_g,E,heat), theme(t\_g,E,X),   \\\hspace{2.8cm} 
              t\_start(t\_g,E,T1), t\_end(t\_g,E,T2), \\ \hspace{2.9cm}temp(T1,X,Tm1), temp(T2,X,Tm2).} \\
C6 & \makecell[l]{0.9::temp(T2,X,TmpD):-isa(t\_g, E, heat), theme(t\_g, E, X),   \\\hspace{2.4cm} 
              t\_end(t\_g,E,T2), dest\_attr(t\_g,E,TmpD).} \\              
               

%

\hline

\rowcolor{Gray}
\hline 
\multicolumn{2}{|l|}{Example queries (with background knowledge)} \\ 
\hline

Q1 & \makecell[l]{isa(enter1, enter\_phy), theme(enter1, air), t\_start(enter1,t1), \\ t\_during(enter1,t2), t\_end(enter1,t3), dest(enter1,engine),\\ channel(enter1,intake). \textit{\%air enters engine through intake}\\ \textcolor{orange}{location(t1,air,?,?). location(t2,air,?,?). location(t3,air,?,?)} }\\

Q2 & \makecell[l]{Same as Q1 $+$ source(enter1, propeller). \textcolor{orange}{location(t1,air,?,?).}}\\

Q3 & \makecell[l]{isa(heat1, heat). theme(heat, water). t\_start(heat1,t1). \\ t\_end(heat1,t3), temp(t1,water,tmp1), dest\_attr(heat1,tdest).\\value(tdest,30,cels). \textcolor{orange}{temp(t3, heat1, X), more\_than(X,t1)?} }\\
\rowcolor{Gray}
\hline 
\multicolumn{2}{|l|}{Inferences} \\ 
\hline
I1 & \makecell[l]{1.0 location(t1,air,engine,out), 1.0 location(t2,air,intake,at).\\ 1.0 location(t3,air,engine,in).} \\

I2 & \makecell[l]{1.0 location(t1,air,propeller,at),  location(t1,air,engine,out).} \\

I3 & 0.9 temp(t3,heat1,X=tdest), more\_than(X=tdest,t1)=True \\
\hline

\end{tabular}\caption{\textbf{Event schemas with temporal implications} of what happens before or after to filled semantic roles shown for \textit{enter} and \textit{heat}.}\label{table:approach:event_temporal}
\end{table}

\textbf{Representing event temporal implications (verb schemas).}
To represent event expectations or temporal implications (such as what happened before, during or after), or more formally, verb schemas~\cite{ferretti2001integrating}, we utilize the \textit{time\_point} argument as introduced earlier. Similar to before, the temporal inferences can differ based on provided situational or semantic role knowledge for the event. As shown in table \ref{table:approach:event_temporal}, C1-C4 represent temporal implications for the event \textit{enter\_phy} (e.g. `the man entered the room'). Here, the additional roles of \textit{source}, \textit{destination} and \textit{channel} are utilized specifying where an object is originating from (source), where it is entering into (destination) and what it passes through (channel) in the process. Further, temporal attributes \textit{t\_start}, \textit{t\_during} and \textit{t\_end} specify the time points when an event starts, when it is going on and when it ends respectively.

C1 thus represents the inference that if \textit{X} is a theme of \textit{E} which is an instance of event \textit{enter\_phy} whose \textit{destination} is stated to be \textit{D}, then at the end of the event, the location of \textit{X} is in \textit{D}. Similarly, C2 and C3 capture locative implications of the theme \textit{X} before or while the event is taking place. C4 additional represents the inference that an object \textit{X} before entering into \textit{D}, is outside \textit{D}. Q1 details the scenario \textit{``air enters an engine through intake"} in an approprite logical/semantic-role representation with \textit{enter\_1} being an instance of \textit{enter\_phy}, its theme being the entity \textit{air}, its destination being \textit{engine} and its channel being \textit{intake}. \textit{t1, t2 and t3} represent the time indicators specific to the instance \textit{enter\_1}. Based on this scenario Q1 queries the locations of air before, during and after the event, and based on application of C2-C4 infers them. Q2 represents a similar scenario but with knowledge that the source was a propeller (``\textit{air enters the engine from the propellor through the intake}"). This enables it to make the additional inference that location prior to \textit{enter} was at the \textit{propellor} as in I2. 

Similarly, C5 and C6 represent temporal implications of the event \textit{heat} with C5 indicating that the object being heated (theme) will have a higher temperature after heating than it had prior to heating. C6 indicates that the temperature after heating will be likely equivalent to specified `dest\_attr' (e.g. \textit{50 degrees celsius} in \textit{``Heat the batter to 50 degrees celsius"}). Q3 details a scenario of water being heated to 30 degree celsius and queries the temperature after heating and whether it will be higher than it was at start of heating.





\textbf{\\Physical and comparative inferences.} 
To perform physical and comparative inferences, we build upon Problog's inbuilt numerical operators (e.g. $>, >=, <$, etc.). As shown in table \ref{table:approach:physics_and_comparitive}, F2 and F4 represent boiling points of water (F1) and olive oil (F3) respectively. C1 denotes the `more\_than' predicate inference which is derived by doing the appropriate $>$ operation on the \textbf{\textit{value}} of its arguments X and Y. 

Using this, physical and comparative inferences can be represented such as in C2 (\textit{``if the empty volume of X is more than size of Y, then X can contain Y"}) and C3 (\textit{``if X is a liquid and its temperature is more than its boiling point, then X likely is boiling"}). Q1 represents a scenario wherein \textit{ballx} has size of $30 cm^{3}$ and \textit{cup1} has empty volume of $20 cm^{3}$, and asks whether \textit{cup1} can contain \textit{ballx}. Based on C2 and C1, this evaluates to False. Similarly Q2 represents a scenario wherein a liquid is being heated to \textit{120 degrees celsius}, and first queries the associated event for the liquid, if the liquid is \textit{water}, and then queries the same for if the liquid is \textit{olive oil}. In the first case, the inference that associated event of \textit{water} is \textit{boiling} is made (due to F1, C1 and C3 -- the heating temperature is higher than boiling point of water) and in the second case no inference can be made.  

This scheme can also support more complex knowledge such as physics laws. C4-C6 provide an example of the ``ideal gas law". C4 captures the implication of proportionality -- \textit{if \{X and Y are proportional and Y increases\}, or \{X and Y are inversely proportional and Y decreases\} then X increases}. C5 represents similarly for X decreasing depending on antecedents. Finally, C6 represents that if volume \textit{V} of a gas \textit{X} is constant, then its temperature \textit{T} and pressure
\textit{P} are proportional. Further rules for other combinations of \textit{P, V} and \textit{T} can be similarly represented. Q3 represents a scenario wherein the \textit{volume} of \textit{air} is known to be constant and its \textit{temperature} is stated to be decreasing (which could be also implicitly provided by stating that \textit{air} is a `theme' of event \textit{cool}). It queries whether the \textit{pressure} decreases, and as shown in I3 evaluates to True based on application of C6 and C5.

\begin{table}[h]
\small
\setlength\tabcolsep{2pt} 
\begin{tabular}{|wl{0.042\columnwidth}|wl{1\columnwidth}|}
\rowcolor{Gray}
\hline
\multicolumn{2}{|l|}{Knowledge} \\ 
\hline
F1 & boiling\_point(water,bp\_water).\\
F2 & value(bp\_water, 100, celsius). \\

F3 & boiling\_point(olive\_oil, bp\_ooil).\\
F4 & value(bp\_ooil, 300, celsius). \\
C1 & more\_than(X,Y):- value(X,Vx,U), value(X,Vy,U), Vx$>$Vy \\
C2 & \makecell[l]{can(X,contain,Y):- empty\_vol(X,Sx), size(Y,Sy), \\\hspace{2.4cm}more\_than(Sx,Sy).} \\
C3 & \makecell[l]{assoc\_event(X,boil):- isa(X,liquid), temp(X,T),\\ \hspace{2cm} boiling\_point(X,Tbp), more\_than(T,Tbp).} \\
C4 & \makecell[l]{increases(X):- (proportional(X,Y), increases(Y)); \\ \hspace{2cm}(invproportional(X,Y), decreases(Y)).} \\  
C5 & decreases(X):- (invproportional(X,Y), increases(Y)); \{..\} \\
C6 & \makecell[l]{proportional(T,P):- isa(X,gas), temp(X,T), vol(X,V),\\pressure(X,P), constant(V). \textit{\%gas law (other combns. omitted)}}  \\

%

\hline

\rowcolor{Gray}
\hline 
\multicolumn{2}{|l|}{Example queries (with background knowledge)} \\ 
\hline
Q1 & \makecell[l]{size(ballx, s1), empty\_vol(cup1, s2), val(s1,30,cm$^{3}$),\\ val(s2,20,cm$^{3}$). \textcolor{orange}{ can(cup1,contain,ball1)?} }\\
Q2 & \makecell[l]{isa(heat1, heat), dest\_attr(heat1,tdest), value(tdest,120,cels..) \\ theme(heat1, water). \textcolor{orange}{assoc\_event(water,E?)?} \\ theme(heat1, olive\_oil). \textcolor{orange}{assoc\_event(olive\_oil,E?)? }}\\
Q3 & \makecell[l]{isa(air,gas), pressure(air,p1), vol(air,v1), temp(air,t1).\\ constant(v1), decreases(t1). \textcolor{orange}{decreases(p1)?}}\\
\rowcolor{Gray}
\hline 
\multicolumn{2}{|l|}{Inferences} \\ 
\hline
I1 & 0.0 can(cup1,contain,ballx) \{\textit{False}\}\\
I2 & \makecell[l]{1.0 assoc\_event(water, E=boil).\\
                  0.0 assoc\_event(oil, E=?). \{\textit{unknown}\}
                }\\
I3 & 1.0 decreases(p1). \{\textit{True}\}\\
\hline

\end{tabular}\caption{\textbf{Physical and comparative inferences}. C1-C3 represent clauses that perform numerical comparisons to perform inferences such as whether an item can contain another object (C2) or whether a liquid will boil at a particular temperature (C3) with example inferences I1 and I2. C4-C6 encode the gas law and capture notion of proportionality/inverse proportionality  with example inference I3. }\label{table:approach:physics_and_comparitive}
\end{table}











\textbf{Higher-order predicate inferences.} Certain inferences and beliefs require higher-order logic wherein a predicate is nested in another predicate. This is especially useful for \textit{sentient} type predicates (e.g. \textit{believes}, \textit{desires}, \textit{prefers}, etc) where the second argument may itself be a predicate/belief, e.g. \textit{believes(student, origin(universe, ``big bang"))} indicating that a student believes the origin of the universe is from the ``big bang". 

Similarly, as shown in table \ref{table:approach:higher_order}, the implication of a person being a vegetarian is captured in C2. It states that if an \textit{person} is \textit{vegetarian}, then they would \textbf{prefer} eating items \textit{F} (being the theme) that are not made of \textit{animal}. Note, this is different from what they \textit{can} physically eat from an anatomical perspective (by virtue of being an \textit{animal}), which is captured in C1 (which states that if \textit{X} is an \textit{animal} and \textit{Y} is a type of \textit{food}, then \textit{X} can probably eat \textit{Y}). Q1 represents a scenario wherein \textit{p1} is a vegetarian, and \textit{p2} is unspecified. Different food items including \textit{kebab}, \textit{tofu} and \textit{pizza} are indicated to be made of different items. When queried what all items \textit{p1} and \textit{p2} \textbf{can} eat, all items are returned as they are all types of food. However, when queried what all items \textit{p1} \textbf{prefers} to eat, the returned items are \textit{tofu} and \textit{pizza} (with 0.5 likelihood due to uncertainty of whether it is made of meat). For \textit{p2}'s case however, all items are returned as possible preferences. 

A similar example is also provided in C3 to encode the inference whether an agent (denoted by \textit{B}) can buy an object (denoted by \textit{Z}). As shown in the antecedents, this inference is true if the owner of \textit{Z} (denoted by \textit{S}) desires something that \textit{B} owns, and also \textbf{believes} the value of that is more than equal to the value of \textit{Z}. Q2 and I2 show different owners of objects, and how only \textit{p3} can buy \textit{home1} as its owner (\textit{p1}) wants a car and \textbf{believes} that the value of \textit{p3}'s car exceeds value of \textit{home1}. Such a scheme could thus be potentially useful for capturing theory of mind and relevant social commonsense.



\begin{table}[h]
\small
\setlength\tabcolsep{2pt} 
\begin{tabular}{|wl{0.05\columnwidth}|wl{0.95\columnwidth}|}
\rowcolor{Gray}
\hline
\multicolumn{2}{|l|}{Knowledge} \\ 
\hline
F1 & 0.7::believes(student, origin(universe, ``big bang")) \\
C1 & \makecell[l]{ \textit{\% Below captures X can eat Y if X animal and Y food}\\ 0.7::can(X, E, Y):- isa(X,animal), isa(Y, food), isa(E,eat).}\\
C2 & \makecell[l]{ \textit{\% Below is X \textbf{prefers} eating Y not from animal if X is veg.}\\ prefers(X, theme(E,F)):- isa(X, person), isa(X,vegetarian), \\isa(E,eat), isa(F, food), agent(E,X), not(made\_of(F,animal)).}\\
C3 & \makecell[l]{ \textit{\% Below is X can buy Z depending on Z \textbf{owner's valuation}.}\\ can(B, buy, Z):- owns(B,C), owns(S,Z), wants(S,C), \\ believes(S, val(C,VC)), believes(S, val(Z,VZ)), \\  more\_than\_equal(VC,VZ). \textit{\%val denotes value}} \\

\hline

\rowcolor{Gray}
\hline 
\multicolumn{2}{|l|}{Example queries (with background knowledge)} \\ 
\hline
Q1 & \makecell[l]{isa(p1, animal), isa(p1, vegetarian), isa(p2, animal),\\ made\_of(kebab,animal), made\_of(tofu, soy),\\   0.5::made\_of(pizza, animal). \\ agent(eat1,p1). \textcolor{orange}{can(p1,eat,F)? prefers(p1,theme(eat1,F))?}
\\ agent(eat2,p2). \textcolor{orange}{prefers(p2,theme(eat2,F))? }}\\
Q2 & \makecell[l]{owns(p1, home1), wants(p1, car), owns(p2, car2),\\owns(p2, home2), owns(p3, car3), \\ believes(p1, val(home1, 30k)), believes(p1,val(car2,20k)), \\believes(p1,val(home2,50k), believes(p1,val(car3,40k).)\\  \textcolor{orange}{can(p2, buy, home1)? can(p3, buy, home1)?}}\\
\rowcolor{Gray}
\hline 
\multicolumn{2}{|l|}{Inferences} \\ 
\hline
I1 & \makecell[l]{1.0 can(p1,eat,F=\{meat,kebab,tofu,pizza\}) \\
                  1.0 prefers(p1,theme(eat1,F=tofu)), 0.5 prefers(.., F=pizza) \\
                  1.0 prefers(p2,theme(eat2,F=\{meat,kebab,tofu,pizza\})).
                 }\\
I2 & \makecell[l]{0.0 can(p2,buy,home1). \\
                  1.0 can(p3,buy, home1). 
                }\\
\hline

\end{tabular}\caption{\textbf{Higher-order inferences.} F1 represents a simple higher-order fact wherein \textit{origin}(..) is an argument of the \textit{belief} of a student. C2 encodes the implication of a person being vegetarian and differentiates the higher-order predicate \textit{prefers} from first-order predicate \textit{can} (as shown in Q1 and I1). C3 represents usage of higher-order predicate \textit{believes} for inferring whether a person can buy an item.}\label{table:approach:higher_order}
\end{table}

\section{Knowledge-base construction and applications}
\label{sec:applications}
In this section, we detail how we extend PrimeNet with our proposed knowledge representation scheme through crowdsourcing and manual annotation. We then illustrate its application for passage-based semantic parsing and question-answering. Our target domain for this exercise was aerospace documents and concepts (due to project funding requirements) with a focus on interpretable and `deeper' semantic inferences for passage-based parsing. 

\subsection{Knowledge collection and analysis}
A total of 574 concepts comprising original ontological nodes, nouns and verbs were identified from frequency analysis of word usage in domain text. A total of ten annotators (including the paper authors) were involved in identifying applicable concept groups within the ontology and extending the ontology with new concept groups, lower-level concepts and relevant predicates. Crowdsourcing was used to collect factual knowledge on type-specific predicates (relations and attributes) associated with each concept and to estimate their probability.  Manual annotation was used for knowledge clauses and more domain-specific facts.

\textbf{Crowdsourcing setup.} 100 concepts were chosen to develop and evaluate the crowdsourcing framework. A survey-based design was used for this. Each concept was first illustrated with an example sentence to specify the word sense in which the concept is to be used.  This was followed by questions to collect knowledge on relations and attributes for that concept requiring responses in the form of free-text phrases, or by selecting applicable options from a list (for knowledge that can be mapped to existing grounded concepts). Participants were recruited via amazon mechanical turk across eight different countries. To ensure high quality of responses, participants first had to pass a word sense selection test involving 3 concepts, wherein for each concept they were shown a sentence using a particular sense of the concept and had to identify the correct sense from four possible options. Only participants who perfectly completed the selection test were permitted to proceed with the survey. Participants were then familiarized with the survey setup by doing a small practice run.  For the actual survey, each participant was asked to annotate 3 concepts and each concept was assigned randomly to at least 3 annotators.  Participants were given 10 minutes for each a concept, after which the survey would automatically move on to the next concept to be annotated. Example UIs for the first round of crowdsourcing are provided in appendix fig. \ref{fig:appendix:cwdsr}.  IRB approval was obtained for the crowdsourcing exercise.

\begin{figure}[t]
\includegraphics[width=0.99\columnwidth]{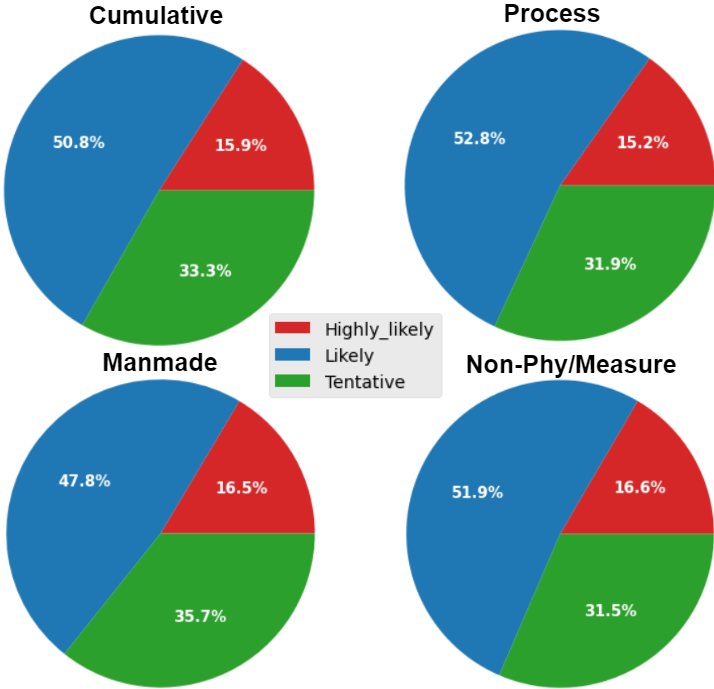}
\caption{\textbf{Crowdsourced certainty levels of facts} for salient concept groups. At least 3 annotators rated each fact and average score was used to filter and identify discrete certainty levels.} 
\label{fig:application:preddistr}
\end{figure}

Free-text responses in the collected knowledge was first mapped to the grounded concepts in the existing knowledge-base through name retrieval and manual checking.  In cases where no matching concepts were found, free-text responses were left as is. The resultant knowledge was then utilized for a second round of crowdsourcing in which responders were asked to rate a knowledge fact on a 5-point scale between \{\textit{Highly disagree}, \textit{Disagree}, \textit{Neutral}, \textit{Agree}, and \textit{Highly Agree}\}. Again, at least three participants were required to rate each knowledge fact; each participant rated 30 facts.  As before, participants were familiarized with the task through examples. An average score for each fact indicating possibility of it being true was computed based on mapping the responses to a range of -2 to 2 (-2 corresponding to highly disagree and 2 corresponding to highly agree). Finally, any fact having score less than equal to 0 was discarded, while scores ranging from 0.0-0.7 were counted as `tentative', 0.7-1.4 as `likely' and  1.4-2.0 as `strongly likely'. The collected responses for salient conceptual types are specified in fig. \ref{fig:application:preddistr}. As shown, a higher percentage of facts across concept types were found to be rated as likely, followed by tentative and highly likely. Concepts of type \textit{Manmade} were found to have a generally lesser percentage (47.8\%) of likely facts in comparison to \textit{Process} and \textit{Non-physical or Measure} types. 

\begin{figure}[t]
\includegraphics[width=0.99\columnwidth]{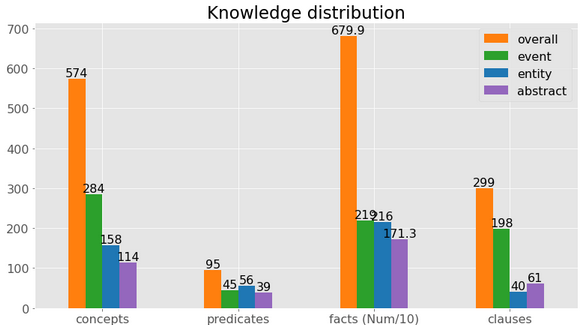}
\caption{\textbf{Knowledge distribution} specifying number of concepts, predicate, facts (divided by 10 in fig. for scale) and clauses amongst event, entity, abstract and all-together categories.} 
\label{fig:application:distrb_of_knowledge}
\end{figure}

\begin{figure*}[t]
\includegraphics[width=0.99\columnwidth]{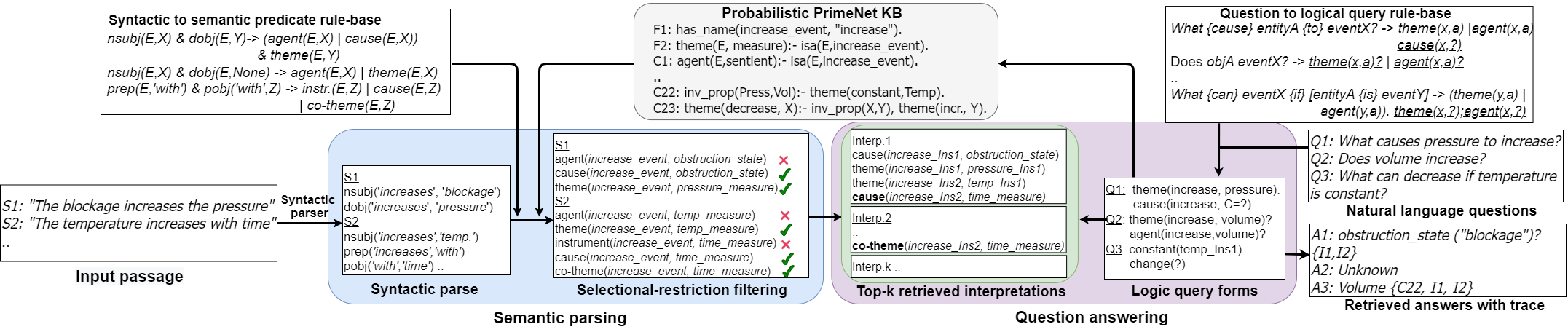}
\caption{\textbf{Utilization of knowledge-base for semantic parsing and question answering}. Input sentences are mapped to possible knowledge-base semantic forms through a syntactic dependency to semantic predicate mapping rule-base. Possible forms are filtered through selectional restriction checking and top-k valid interpretations are maintained. Question answering is similarly performed by converting the question into valid logical queries through a template rule-base, which are then executed against the knowledge-base as well as text-parsed facts to retrieve answers.} 

\label{fig:application:semantic_parsing_and_QA}
\end{figure*}

\textbf{Manual annotation.} To annotate clauses and more domain-specific facts, manual annotation was performed by 6 annotators having knowledge of the domain. Annotation was performed primarily to identify salient physical properties of entities, part-whole connections, event schemas, inter-event relations, entity-event relations and physics laws. Each annotated knowledge was checked by at least one other annotator for quality checking, and the final knowledge was added into the knowledge-base in form of concept facts as well as clauses. 

Fig. \ref{fig:application:distrb_of_knowledge} displays the resultant knowledge base statistics. As shown, it consists of 284 events, 158 real entities and 114 abstract concepts. A total of 95 predicates are used with 45 being broadly event-typed predicates, 56 being real world entity-typed and 39 being abstract-typed. Note, that predicates amongst groups overlap based on inheritance from top-level nodes. Cumulatively across concepts, the knowledge-base encodes a total of 6799 facts (of which 2190 are for events and 2160 are for real entities) and 299 clauses (of which 198 are for events).




\subsection{Knowledge-based semantic parsing and question answering}
The resultant knowledge-base was applied for passage based semantic parsing and question answering. Passages were obtained from chapters of aerospace manuals in the target domain. As shown in fig. \ref{fig:application:semantic_parsing_and_QA}, here semantic parsing involved converting the input text passage into logical forms that have the same representation format as facts in our knowledge-base. The predicates for the target logical form are the same as the ones in our knowledge-base. Free-text phrases are mapped to appropriate grounded knowledge-base concepts through word-sense disambiguation (described later) or left as is (in case no mapping concept was found). For example, as shown in the figure, the free-text phrase ``blockage" in the first sentence retrieves the concept \textit{obstruction\_state}, which is thereafter used in logic semantic forms.

\textbf{Semantic parsing module.} While one could apply a deep-learning based semantic parser~\cite{dong2018coarse,campar} for this task, we currently developed a rule-base parser as we did not have a significant amount of annotated logic forms corresponding to input texts to utilize as training data. We developed a rule-base that maps syntactic dependency antecedents to applicable semantic forms (with predicates from our knowledge-base). This mapping is many-to-many, i.e. a syntactic relation combination can map to multiple semantic predicates while the same semantic predicate can be activated by multiple syntactic relation combinations. Hence, it can lead to multiple possible interpretations as shown in fig. \ref{fig:application:semantic_parsing_and_QA}, wherein the first sentence \textit{``The blockage increases the pressure"} is mapped to two mutually-exclusive facts \textit{cause(increase, obstruction)} and \textit{agent(increase, obstruction)}. To resolve this, we perform selectional restriction filtering, wherein for a given event role (or more generally predicate), the applicable conceptual types for the filled argument are specified. In interpretations wherein the filled argument does not satisfy the applicable conceptual types, the interpretation is removed or its likelihood is reduced. An example is given in C1 of the same figure's knowledge-base which states that the agent of \textit{increase\_event} is a type of \textit{sentient}. Since this fails for \textit{agent(increase, obstruction)}, this possible mapping is discarded. However, in the case of the second sentence, \textit{``The temperature increases with time"}, for the two mutually-exclusive mappings \textit{cause(increase, time)} and \textit{co-theme(increase, time)}, both cases satisfy selectional restriction checks and hence are stored as two distinct interpretations. 

Distinct interpretations are also maintained in cases wherein multiple concepts are retrieved for a given text phrase (during word sense disambiguation) and satisfy further selectional restriction checks. Note that after finding matching concepts, an `instance' of the concept is created (e.g. \textit{increase\_Ins1} corresponds to an instance of event \textit{increase} that was specified by text-phrase \textit{``increases"}, similarly, \textit{pressure\_Ins1} corresponds to an instance of \textit{pressure}). An instance is re-used for future sentences if no determiner is provided, and a new instance for a referred concept is only created if the determiners are in \textit{\{a, another, an\}}. This process is repeated for all sentences of the given passage, and the top-k interpretations are stored after each rule application (for our work, we use k=3). For each passage, the parsed forms are reset, i.e. passages are parsed independently, and we currently do not perform discourse linking. 

Overall, 49 rules were identified based on domain text for mapping syntactic dependency forms into semantic forms (corresponding to knowledge-base fact representations). The spacy parser~\cite{spacy2} was utilized to obtain dependency trees for sentences. Further, for anaphora and coreference resolution, we used the \textit{``neuralcoref"} package within spacy, and will look into developing appropriate knowledge-based methods in future work. The resultant parser was applied on 267 sentences across 21 passages.

\textbf{Question answering module.}
Given the parsed semantic forms and existing knowledge in the knowledge-base, logical queries can be directly performed as previously illustrated in sec. \ref{sec:approach}. The different parsed interpretations are specified as different knowledge sources. Hence, when the inference is performed, the source trace specifies which interpretation was used to derive the answer in addition to the originally encoded knowledge in the knowledge-base. To allow language questions to be queried, we developed an additional template-based rule-base to convert the question into appropriate logical queries. As shown in fig. \ref{fig:application:semantic_parsing_and_QA}, the question to logical query rule-base identifies phrases matching to predicates(such as \textit{cause} and \textit{to} in \textit{What causes A to X?} which are mapped to event assertions \textit{theme} or \textit{agent}, and the query \textit{cause(X,?)}). 

While some of the queries can be directly answered through the text-parsed knowledge (as for \textit{What causes pressure to increase?}), others may require inference from both text-parsed knowledge and the knowledge-base (as in \textit{What can decrease if the temperature is constant?}). In both cases, a trace for the answer is provided specifying which knowledge facts and clauses were utilized in answering the query, thereby providing clear interpretability in answering the question. Further, likelihoods for different possible answers can be derived based on the activated probabilistic facts or clauses, and in cases where no applicable knowledge can be found, the \textit{unknown} answer is generated. 

Overall, the resultant knowledge-based system allows for more interpretable semantic parsing and question answering; however, the input sentence and question processing capabilities are currently constrained to the rule-base and will benefit from integration with neural methods in future work. Finally, the described system can also be used to perform non-monotonic reasoning or make defeasible inferences, wherein inferences and their likelihoods may progressively change based on new evidence (or facts) provided by the text.

\section{Conclusion}
\label{sec:conclusions}
We introduced a probabilistic-logic based commonsense representation framework to encode a larger variety of world knowledge and to represent conceptual beliefs with varying likelihoods. We also proposed a hierarchical conceptual ontology designed to identify salient concept relevant relations and enable beliefs to be encoded at different conceptual levels and re-used where applicable through inheritance. We illustrated the knowledge representation scheme and how it can encode different types of commonsense knowledge including contextually-dependent inferences, event schemas / temporal implications, physical and comparitive inferences, and higher-order inferences. We then applied the representation scheme and ontology to extend the PrimeNet knowledge-base by crowdsourcing and manually-encoding knowledge for the aerospace domain. Finally, we illustrated how the resultant knowledge-base can be utilized for more interpretable passage-based semantic parsing and question answering.


\section*{Acknowledgement}
This research is supported by A*STAR under its “K-EMERGE:Knowledge Extraction, Modelling and Explainable Reasoning for General Expertise” programme (Grant number A19E2b009). The crowdsourcing setup was approved by the A*STAR Institutional Review Board (IRB Reference: 2019-016).







\bibliographystyle{elsarticle-num-names} 
\bibliography{cas-refs}

\appendix
\label{sec:appendix}

\section{Probabilistic logic programming primer and relevant syntax}
Table \ref{table:appendix:primer} lists probabilistic logic programming terminology and ProbLog syntax relevant to our work's discussion. An example representation of the natural language belief \textit{`X can move with base likelihood 0.9 if X has a leg or a wheel and X is not in a stuck state'} in logical form is ``\texttt{0.9::can(X,move):- (has(X,leg); has(X,wheel)), not(state(X,stuck)).}" 
For a more comprehensive overview of probabilistic logic programming, we refer the reader to \cite{de2015probabilistic}.

\begin{table*}[t]
\small
\setlength\tabcolsep{1.5pt}
\begin{tabular}{|wc{0.11\columnwidth}|wl{0.49\columnwidth}|wl{0.37\columnwidth}|}
\hline
\rowcolor{Gray}
Term & Description & Examples\\
\hline
Predicate & \makecell{A relation (e.g. isa, causes)  or attribute (e.g. size, color)} & \makecell{isa, shape, related\_to, color..} \\
\hline
Constant & \makecell{A term that is grounded and fixed (e.g. numbers, text-phrases \\ and symbols with lower-case first letter)} & \makecell{`X eats Y', -23, \\ person, `Bob', x} \\
\hline
Variable & \makecell{An ungrounded term that can take values during inference. \\Denoted by symbols with upper-case first character \\} & \makecell{X, Y, Event, \\Bob, Person} \\
\hline
Atom & \makecell{A predicate with n arguments; Syntax: pred(arg0,arg1..)} & \makecell{has\_part(car,wheel,4), isa(X,X)} \\
\hline
\makecell{Logical \\operators} & \makecell{(i) `;' denotes `logical or', (ii) `,' denotes `logical and' \\(iii) 'not' or `$\backslash+$' denotes `not' (iv) `.' denotes end of fact/clause} & \makecell{isa(car,vehicle). not(isa(car,apple)). \\ isa(car,on); isa(car,off).} \\
\hline
\makecell{Probabilistic \\ fact} & \makecell{A declarative belief with a given base likelihood \\ (which if unspecified assumed to be 1)} & \makecell{0.5::can(x,buy,car). isa(X,X).\\0.98::has(person,face). isa(person,physical).} \\

\hline
\makecell{Probabilistic \\ clause or rule} & \makecell{An inferential belief where  a head atom `h' is inferred \\ with a base clause likelihood  $P_c$ from computation of body \\ (or antecedent) set of atoms (denoted by `b') as follows:\\ $P(h)=P_c \times \prod{P(b)}$} & \makecell{$P_c$:: head:- a1, (a2;a3). \textit{(Base syntax)}\\ 0.6::can(X,move):-has(X,leg). \\can(X,buy,Y):-has(X,Z),(value(Z)$>=$value(Y)).} \\
\hline


Model & \makecell{Corresponds to a set of facts and rules representing a \\world `model' on which inferences can be performed} & \makecell{0.5::can(item,move).\\0.9::can(X,move):-(has(X,wheel);has(X,leg)).} \\
\hline

Query & \makecell{An atomic query to infer from  a given world likelihood of a  \\ fact (e.g. can(human,move)?) or valid constants for variables \\ for an atom  (e.g. can(X,move)?) } & \makecell{can(human,move)? $=>$ 0.9  \\ can(X,move)?$=>$ X=item (0.5); \\X= human(0.9) ..} \\
\hline
\end{tabular}\caption{Terms in probabilistic logic programming and ProbLog syntax relevant to our work's discussion.}\label{table:appendix:primer}
\end{table*}

\section{Crowdsourcing annotation UI}
\begin{figure}[h]
\includegraphics[width=0.99\columnwidth]{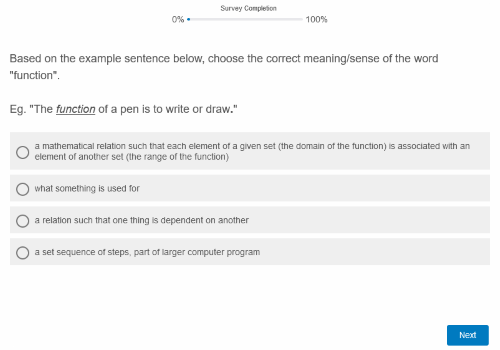}
\caption{Crowdsourcing round 1 screening example} 
\label{fig:appendix:cwdsr}
\end{figure}

\begin{figure}[h]
\includegraphics[width=0.99\columnwidth]{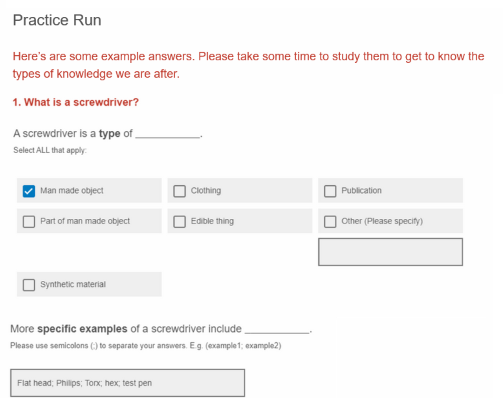}
\caption{Crowdsourcing round 1 practice question example} 
\label{fig:application:knowledge}
\end{figure}


\begin{figure}[h]
\includegraphics[width=0.99\columnwidth]{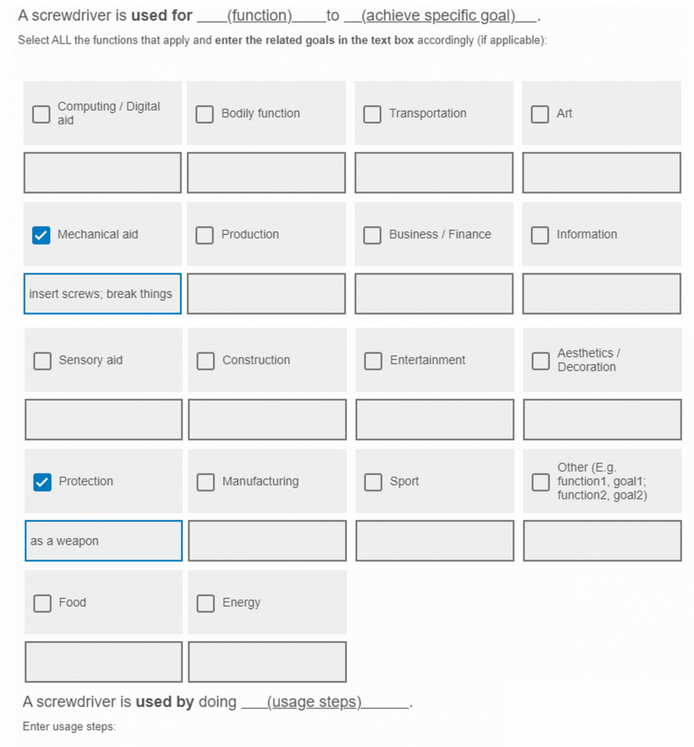}
\caption{Crowdsourcing round 1 main task example} 
\label{fig:application:knowledge}
\end{figure}

\begin{figure}[h]
\includegraphics[width=0.99\columnwidth]{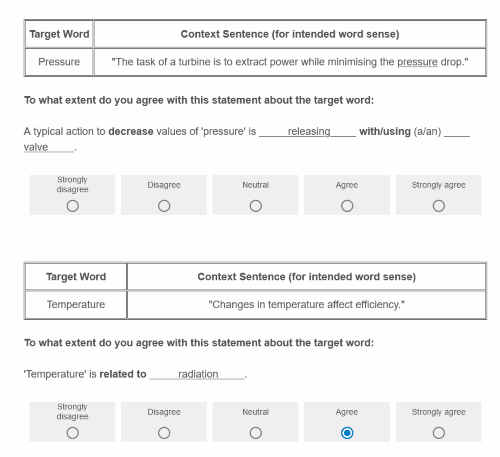}
\caption{Crowdsourcing round 2 main task example} 
\label{fig:application:knowledge}
\end{figure}






\end{document}